%% file: main.tex
\documentclass[10pt,twocolumn,letterpaper]{article}

\usepackage{iccv}              


\usepackage[accsupp]{axessibility}  
\usepackage[dvipsnames]{xcolor}
\usepackage{amsmath}
\usepackage{amssymb}
\usepackage{graphicx}
\usepackage{adjustbox}

\definecolor{iccvblue}{rgb}{0.21,0.49,0.74}
\usepackage[pagebackref,breaklinks,colorlinks,allcolors=iccvblue]{hyperref}

\newcommand{\setNotation}[1]{\mathcal{#1}}

\newcommand{\setImRef}{\setNotation{I_{R}}}
\newcommand{\setImQ}{\setNotation{I_{Q}}}
\newcommand{\setRef}{\setNotation{R}}

\newcommand{\setPoseRef}{\setNotation{P_R}}
\newcommand{\setPoseQ}{\setNotation{P_Q}}
\newcommand{\setMap}{\setNotation{M}}

\newcommand{\smallDG}{\checkmark}
\newcommand{\mediumDG}{\checkmark \checkmark}
\newcommand{\largeDG}{\checkmark \checkmark \checkmark}

\newcommand{\iccvcolor}{\textcolor{black}} 

\def\paperID{*****} 
\def\confName{ICCV}
\def\confYear{2025}

\title{The Overlooked Value of Test-time Reference \iccvcolor{Sets} in Visual Place Recognition}

\author{Mubariz Zaffar\\
ME, TU Delft,\\
The Netherlands.\\
{\tt\small m.zaffar@tudelft.nl}
\and
Liangliang Nan\\
ABE, TU Delft,\\
The Netherlands.\\
{\tt\small liangliang.nan@tudelft.nl}
\and
Sebastian Scherer\\
RI, CMU,\\
USA.\\
{\tt\small basti@andrew.cmu.edu}
\and
Julian F. P. Kooij\\
ME, TU Delft,\\
The Netherlands.\\
{\tt\small j.f.p.kooij@tudelft.nl}
}

\begin{document}
\maketitle
\input{chapters/abstract}

\input{chapters/introduction}

\input{chapters/relatedwork}

\input{chapters/methodology}

\input{chapters/experiments}

\input{chapters/conclusions}

{
    \small
    \bibliographystyle{ieeenat_fullname}
    \bibliography{main}
}

\end{document}

%% file: chapters/abstract.tex
\begin{abstract}
Given a query image, Visual Place Recognition (VPR) is the task of retrieving an image of the same place from a reference database with robustness to viewpoint and appearance changes. Recent works show that some VPR benchmarks are solved by methods using Vision-Foundation-Model backbones and trained on large-scale and diverse VPR-specific datasets. Several benchmarks remain challenging, particularly when the test environments differ significantly from the usual VPR training datasets. We propose a complementary, unexplored source of information to bridge the train-test domain gap, which can further improve the performance of State-of-the-Art (SOTA) VPR methods on such challenging benchmarks.
\iccvcolor{Concretely, we identify that the test-time reference set, the ``map'', contains images and poses of the target domain, and must be available before the test-time query is received in several VPR applications. Therefore, we propose to perform simple Reference-Set-Finetuning (RSF) of VPR models} on the map, boosting the SOTA ($ \approx2.3\%$ increase on average for Recall@1) on these challenging datasets. Finetuned models retain generalization, and RSF works across diverse test datasets.


\end{abstract}

%% file: chapters/introduction.tex
\section{Introduction}
Given a query image and a database of geo-tagged reference images, the task of a Visual Place Recognition (VPR) method is to retrieve from the database a correct matching reference image for this query. What is considered as a correct match is ill-defined, but most VPR benchmarks consider any reference image within a fixed (e.g., 25-meter) circular radius of the query location as a correct match~\cite{berton2022deep}. \iccvcolor{VPR has many applications, such as in landmark retrieval~\cite{weyand2020google}, 3D modeling~\cite{agarwal2011building}, image search~\cite{tolias2016image} and map-based localization~\cite{thoma2019mapping, zhu2018visual}. \textit{These applications of VPR require that the test time reference set (the map) is available offline, i.e., before a test-time query is received.}\footnote{\iccvcolor{We acknowledge that there are other applications of VPR where the reference map may not be available offline, such as in SLAM. These applications are \underline{not} the focus in this work.}}}

\begin{figure}
    \centering
    \includegraphics[width=1.0\linewidth]{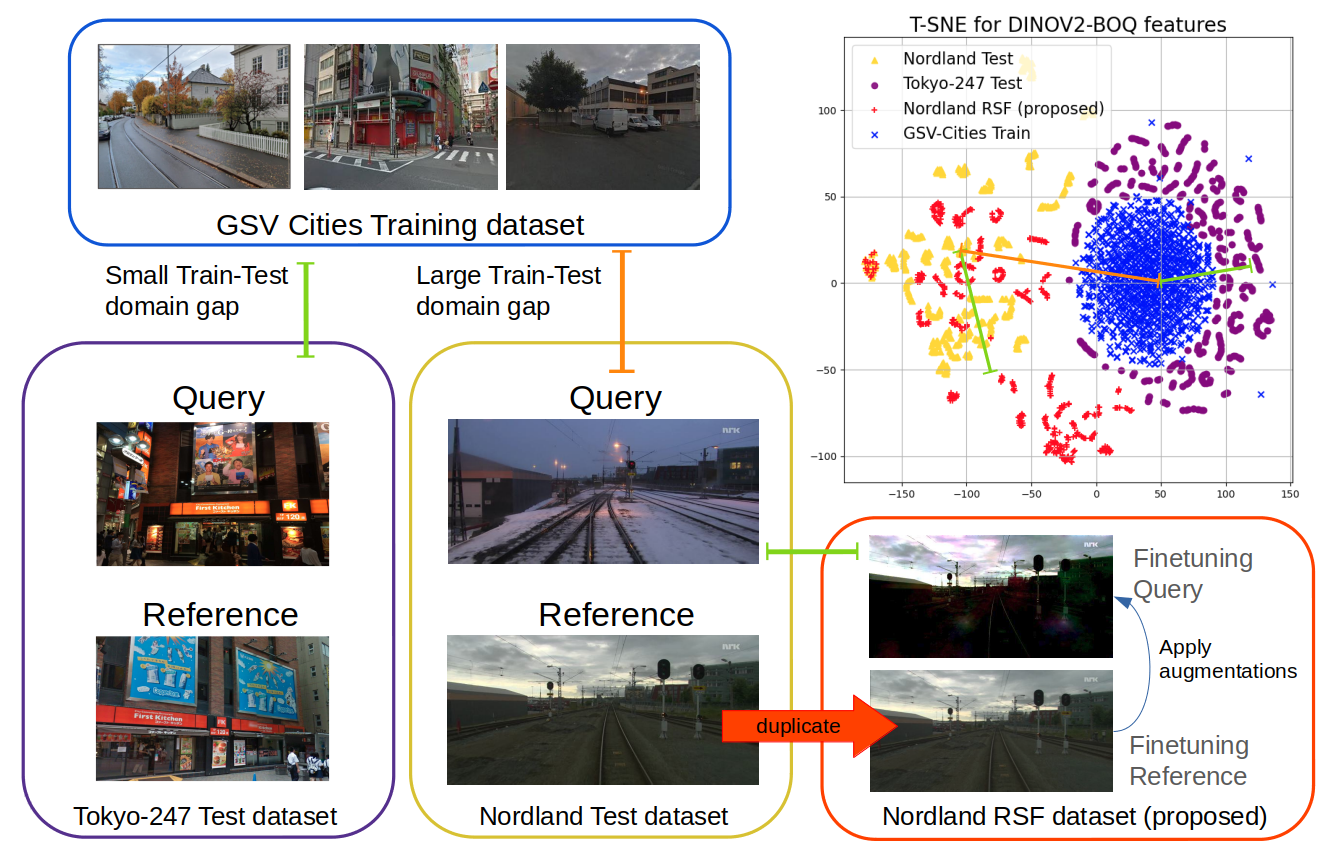}
    \caption{Large-scale VPR training datasets are usually created from Google Street View ~\cite{ali2022gsv}, e.g., the GSV-cities dataset.
    Thus, models trained in these environments perform well (SOTA Recall@5 $\sim98-99\%$) for similar test datasets, e.g., the Tokyo-247 dataset~\cite{arandjelovic2016netvlad}, but suffer in unseen environments, e.g., the railway-tracks of the Nordland dataset~\cite{nordlanddataset}. 
    A train-test domain gap exists, as evident in the T-SNE projection of descriptors computed using BoQ-DinoV2~\cite{ali2024boq} for randomly sampled images of these datasets. Descriptors from the Tokyo-247 dataset form a single cluster with the GSV-cities dataset, while the Nordland dataset is further away. Creating a finetuning dataset by using the freely available test-time reference images could help bridge the train-test domain gap.}
    \label{fig:vprtrainandtest}
\end{figure}

Traditionally, the most investigated challenges in VPR have been viewpoint and appearance changes between the matching query and reference images, the so-called query-ref domain gap~\cite{lowry2015visual}. Thus, the objective of VPR methods is to extract representations robust to these variations. 
Given this objective, VPR benefited significantly through neural networks trained on large-scale VPR-specific datasets~\cite{berton2022deep}. More recently, this has been complemented by adapting strong general-purpose \textit{Vision-Foundation-Model backbones} (VFM) to the task of VPR, e.g., the DinoV2 vision transformer~\cite{oquab2024dinov2, ali2024boq, lu2024cricavpr, izquierdo2024optimal}. As a result, test datasets with large query-ref domain gap (e.g., Tokyo-247~\cite{arandjelovic2016netvlad} and SVOX-night/snow~\cite{berton2021adaptive}) that were previously challenging for VPR methods now seem solved ($\sim98-99\%$ Recall@5) by the State-of-the-Art (SOTA)~\cite{ali2024boq, lu2024cricavpr, izquierdo2024optimal}.

However, another important but less investigated challenge in VPR is the train-test domain gap, i.e., when the test dataset is from a different environment and/or device than the training dataset. It could be hypothesized that SOTA VPR methods would already be robust to this gap, since VFM backbones are known to generalize across datasets and tasks~\cite{awais2025foundation}, and more so, when finetuned on diverse VPR-specific training data~\cite{ali2022gsv}. We examine this hypothesis, revealing that the current SOTA VPR methods still suffer from the train-test domain gap. Details of this will follow later in the section~\ref{sec:standardvprformulation}.

To address the challenge posed by the large train-test domain gap in VPR, we propose a strategy complementary to the typical curation of larger training datasets and/or using stronger VFM backbones. \iccvcolor{A case is made for using the unexplored reference set in test datasets to finetune the SOTA in VPR. We argue that since this reference set with labeled (poses) images is freely available beforehand in various VPR applications and/or could even be obtained online, it is permissible to use it to bridge the train-test domain gap. Thus, outlining the two assumptions made in our work: a) the test-time reference set is available offline, b) there are resources available at test-time to finetune a VPR model.}

Given this argument, we illustrate in Fig.~\ref{fig:vprtrainandtest} the train-test domain gap in VPR. A T-SNE~\cite{van2008visualizing} projection of two VPR test datasets, Tokyo-247~\cite{arandjelovic2016netvlad} and Nordland~\cite{nordlanddataset}, is shown along with the diverse GSV-Cities training dataset. The Tokyo-247 dataset contains urban scenes similar to the GSV-Cities and hence both form a single cluster, while the Nordland dataset contains railway tracks unlike GSV-Cities and forms a separate cluster. Our proposal is simply that the reference set in test datasets (e.g., Nordland) could be combined with image augmentations to create a new finetuning dataset that has a smaller train-test domain gap than the original GSV-Cities dataset. \iccvcolor{Domain knowledge can then be injected into the model using this proposed finetuning dataset, akin to domain adaptation in other computer vision tasks such as classification~\cite{kang2019contrastive}}. 

However, this raises several questions: a) Is finetuning of VFM-based VPR methods on small test datasets useful? b) Do the finetuned models still generalize to other test datasets? c) Can a single finetuning strategy work across diverse test datasets?  We will present a simple self-supervised strategy, namely, Reference-Set-Finetuning (RSF), to answer these questions. 

%% file: chapters/relatedwork.tex
\section{Related Work}
Visual place recognition was first surveyed in the seminal work of Lowry \textit{et al.}~\cite{lowry2015visual}, which coincided well with the rise of deep learning for computer vision. The three most fundamental challenges identified by Lowry \textit{et al.} in VPR are matching images given viewpoint changes, appearance changes due to illumination, seasons, dynamic objects, etc., and perceptual-aliasing. For handling viewpoint and appearance changes, VPR requires robust image representations, and thus this formulation of VPR as a (deep) representation learning problem led to many works that achieved state-of-the-art VPR performance under challenging conditions~\cite{arandjelovic2016netvlad, radenovic2018fine, revaud2019learning, khaliq2019holistic, leyva2021generalized, berton2022rethinking, wang2022transvpr, ali2023mixvpr, keetha2023anyloc, zaffar2021vpr}. 


Deep-learning-based VPR methods can be broadly categorized based on their underlying novelty, such as the use of a novel loss function~\cite{revaud2019learning, leyva2021generalized, thoma2020geometrically}, better training data~\cite{berton2022rethinking, ali2022gsv}, new architectures~\cite{wang2022transvpr, zhang2021vector, yu2019spatial}, different data augmentations~\cite{musallam2024self, jang2023study, chen2024self}, and new methods for feature aggregation~\cite{arandjelovic2016netvlad, radenovic2018fine, hausler2021patch,ali2023mixvpr}. The work of Berton \textit{et al.}~\cite{berton2022deep} recently created a model zoo based on different combinations of the aforementioned key modules of a VPR system, which is freely accessible online. Since deep learning mainly benefits from larger training datasets, a number of training datasets have been proposed and used in VPR, e.g., the Pitts-250k dataset~\cite{arandjelovic2016netvlad}, Mapillary Street Level Sequences dataset~\cite{warburg2020mapillary}, San-francisco-XL~\cite{berton2022rethinking} dataset, or the GSV-Cities dataset~\cite{ali2022gsv}. The use of vision-transformers in VPR was first studied in TransVPR~\cite{wang2022transvpr}, where image features are first extracted using a CNN, and then a transformer encoder is used to aggregate these features into a global descriptor. This work was followed up by R2former~\cite{zhu2023r2former}, where a vision transformer is used for both retrieval and re-ranking, and operates directly on image patches. 

However, machine-learning-based methods are known to suffer from train-test domain gap~\cite{ben2006analysis}, and such is the case for VPR too, as observed in existing benchmarks~\cite{zaffar2021vpr, berton2022deep}. To bridge this gap, a standard use of the test time reference set in the VPR community is query expansion, where first, an initial retrieval is performed, followed by spatial verification, and then selection of the top-ranked spatially-verified images to form a single or multiple \textit{expanded} queries used for a secondary retrieval~\cite{chum2007total, chum2011total}. The reference set has also been used to create test-time specialized vocabularies for feature aggregators, such as VLAD~\cite{arandjelovic2013all, keetha2023anyloc}. Other than achieving better retrieval performance, the test-time reference set has also been used to estimate uncertainty in VPR~\cite{zaffar2024estimation}. All of these methods work with off-the-shelf feature extractors and do not aim to finetune the feature extraction module given the test-time reference set.

A direction other than train-test domain adaptation is to learn universal and generalizable feature extractors. This has gained significant traction recently, after the release of the DinoV2~\cite{oquab2024dinov2} Vision-Foundation Model (VFM), and other such foundation models~\cite{radford2021learning, kirillov2023segment}. Anyloc~\cite{keetha2023anyloc} investigated using DinoV2 as an off-the-shelf feature extractor. Many concurrent works subsequently showed that the performance benefits are significantly larger when DinoV2 is finetuned on VPR-specific data and training objectives~\cite{ali2024boq, lu2024cricavpr, izquierdo2024optimal, lu2024towards}.  CricaVPR~\cite{lu2024cricavpr} proposes to use correlation between images in the batch with feature aggregation at multiple scales to produce robust global features. SALAD~\cite{izquierdo2024optimal} uses the Sinkhorn algorithm to aggregate the global and local DinoV2 tokens for VPR. Authors of SelaVPR~\cite{lu2024towards} add serial and parallel adapters to the DinoV2 architecture. Finally, BoQ~\cite{ali2024boq} proposes to learn queries useful for VPR using the attention mechanism of transformers, and demonstrates that these learnable queries work with both older (ResNet) and newer (DinoV2) feature extraction backbones. 

These methods collectively show that VFMs (e.g., DinoV2) have directly benefited the VPR community and that stronger backbones, i.e., larger models trained on larger datasets, can directly improve VPR. However, we report that some VPR benchmarks, with a large train-test domain gap, still remain unsolved. In this context, the contributions of our work are as follows:

\begin{itemize}
    \item \iccvcolor{Our comparison of concurrent VFM-based SOTA VPR methods reveals that these methods suffer from a train-test domain gap. It is demonstrated that the freely available test-time reference set can be used to extract useful domain knowledge for VPR applications where the reference map is available offline. }

    \item \iccvcolor{A simple Reference-Set-Finetuning (RSF) strategy is proposed to address the train-test domain gap for such VPR applications. The proposed finetuning improves the SOTA in VPR, and the RSF models retain generalization to other test datasets. RSF works across diverse datasets and is compatible with different VPR methods.}
\end{itemize}



%% file: chapters/methodology.tex
\section{Methodology}
We first formalize VPR, then formulate the use of deep learning in VPR, and finally describe the RSF strategy proposed in this work. 

\subsection{Formalizing VPR}
The goal of VPR is to find one or multiple reference images $I_i \in \setImRef$ that match the place of a query image $I_q \in \setImQ$ given a set of reference images $\setImRef$ with known poses $\setPoseRef$. The pose of $I_q$ is then approximated by the pose of its nearest neighbour references in $\setImRef$. In its standard formulation, VPR consists of an offline map preparation stage and an online retrieval stage. The unknown pose $p_q$ for the query $I_q$ can then be approximated from the poses of the matched references $p_i \in \setPoseRef$~\cite{pion2020benchmarking}.


In the offline phase, 
a VPR method $G$ is applied to every reference image $I_i \in \setImRef$
to obtain $D$-dimensional reference feature descriptors $f_i = G(I_i)$.
The method $G$ is usually a trained neural network~\cite{masone2021survey} or a handcrafted feature descriptor~\cite{dalal2005histograms}.
The resulting VPR map $\setMap = (\setImRef , \setRef,\setPoseRef)$
contains the reference feature descriptors set $\setRef = \{ f_1, \cdots f_N \}$, where each descriptor $f_i$ is associated with a corresponding pose $p_i \in \setPoseRef$.

In the online retrieval stage, the same method $G$ is applied to the query image $I_q$, and its descriptor $f_q = G(I_q)$ is compared to the reference descriptors in the map $\setMap$.
This can be achieved through an efficient $K$-nearest neighbor lookup,
considering the L2-distances $d_i = || f_{i} - f_q ||_2$ between each reference $i$ and the query $q$.

\subsection{Relating the current SOTA in VPR to train-test domain gap}
\label{sec:standardvprformulation}
VPR in deep learning is generally formulated either as a representation learning task~\cite{arandjelovic2016netvlad} or a classification~\cite{berton2022rethinking} task. We use the former formulation in this paper. 
A deep-learning-based VPR method $G$ consists of four major choices: a feature extraction backbone $B$, a feature aggregator $P$, a training dataset $D$, and a metric-learning loss function $\mathcal{L}$. The backbone $B$ and aggregator $P$ are compositional and together form the method $G$, such that $f_i = G(I_i) = P(B(I_i))$. This VPR method $G$ is then trained on the training dataset $D$ by minimizing the loss $\mathcal{L}$. The training dataset $D$ is itself composed of four sets, such that $D=(\setImQ^{train}, \setPoseQ^{train}, \setImRef^{train}, \setPoseRef^{train})$, where for every $I_q \in \setImQ$, the true and false matching reference images $I_i$ are defined usually based on the spatial proximity of their corresponding poses in $\setPoseQ^{train}$ and $\setPoseRef^{train}$, respectively, or based on visual overlap~\cite{berton2023eigenplaces}. 

The choice of backbone in VPR is primarily motivated by advances in other vision tasks, and we have thus seen a change from using VGG~\cite{arandjelovic2016netvlad} and ResNet-based backbones~\cite{berton2021viewpoint, ali2023mixvpr} to domain-agnostic Vision-Foundation-Model (VFM) backbones~\cite{keetha2023anyloc, lu2024cricavpr, izquierdo2024optimal, ali2024boq}. 
For a given backbone $B$, different types of aggregators could be trained as $P$, for example, a NetVLAD layer~\cite{arandjelovic2016netvlad}, GeM layer ~\cite{radenovic2018fine}, or the recently proposed Bag-of-learnable-Queries (BoQ)~\cite{ali2024boq}, etc. BoQ has been shown to outperform other aggregators trained on the same dataset with the same backbone~\cite{ali2024boq}.

Once the architecture $G=P(B(I_i))$ is fixed, the training loss $\mathcal{L}$ could be the distance-based loss~\cite{thoma2020geometrically}, relative-pose-based loss~\cite{melekhov2017relative}, triplet loss~\cite{wang2014learning}, or the multi-similarity loss~\cite{wang2019multi}, etc. These losses could be minimized on different training datasets, for example, the Pitts-250k dataset~\cite{arandjelovic2016netvlad}, Mapillary Street Level Sequences dataset~\cite{warburg2020mapillary}, San-francisco-XL~\cite{berton2022rethinking} dataset, or the GSV-Cities dataset~\cite{ali2022gsv}. The purpose of these training datasets is to learn a generalizable feature extractor $G$ that works well in different domains, and thus the training datasets must be as diverse as possible. From existing literature, the GSV-cities dataset~\cite{ali2022gsv} is the most diverse training dataset in VPR.

Provided this formulation, would a VPR method $G$, employing a VFM backbone (e.g., DinoV2) trained on a large-scale diverse VPR dataset (e.g., GSV-Cities) with SOTA aggregation (e.g., BoQ), resolve the train-test domain gap?
We examine this by benchmarking the performance (Recall@5) in Table~\ref{tab:vprontestdatasets} of three DinoV2-based SOTA VPR methods that were published almost simultaneously~\cite{ali2024boq, izquierdo2024optimal, lu2024cricavpr}. All methods are trained on the GSV-cities dataset~\cite{ali2022gsv}: the most diverse training dataset in VPR, containing viewpoint and appearance changes from many streets across the world. The reported performance suggests that the test datasets with small train-test domain gap are almost solved by these SOTA VPR methods, despite their large query-ref domain gap. But some other test datasets, such as Nordland~\cite{nordlanddataset} and AmsterTime~\cite{yildiz2022amstertime} with archival reference images, where the test environments differ significantly from the training dataset, still present a challenge.\footnote{Please note that we do \textit{not} refer to the presence/absence of train-test domain gap in the various VPR test datasets in binary terms, but in a proportional manner. That is, while there is still a train-test domain gap between the GSV-cities dataset and the solved test datasets, this gap is larger for the unsolved datasets.} 

\begin{table*}
    \centering
    \begin{adjustbox}{width=\textwidth}
    \begin{tabular}{c|c|ccccccc|c}
        \hline
         & Backbone & SVOX-Snow & SVOX-Night & Pitts-250k & Tokyo-247 & Nord. & Eyn. & Ams-AR & Avg.\\
         \hline
        Query-Ref gap &  & \mediumDG  & \mediumDG & \mediumDG & \largeDG & \largeDG & \smallDG & \largeDG & \\  
        Train-Test gap &  & \smallDG  & \smallDG & \smallDG & \smallDG & \largeDG & \mediumDG & \largeDG & \\  
        \hline
        MixVPR~\cite{ali2023mixvpr} ('23) & ResNet50 & 98.4  & 79.5 & 98.2 & 91.7 & 86.8 & 93.2 & 60.4 & 88.5 \\
        BoQ~\cite{ali2024boq} ('24) & ResNet50  & 99.5  & 94.7 & 98.5 & 95.9 & 91.1 & 94.9 & 75.4 & 93.8 \\
        \hline
        Crica~\cite{lu2024cricavpr} ('24) & DinoV2 & 99.0  & 95.0 & 99.0 & 97.1 & 96.2 & 94.9 & 83.9 & 95.6\\
        SALAD~\cite{izquierdo2024optimal} ('24) & DinoV2  & 99.7  & 99.3 & 99.1 & 96.8 & 93.5 & 95.0 & 79.7 & 95.4\\
        BoQ~\cite{ali2024boq} ('24) & DinoV2 & \textcolor{green}{99.7}  & \textcolor{green}{99.4} & \textcolor{green}{99.1} & \textcolor{YellowGreen}{97.8} & \textcolor{Goldenrod}{95.9} & \textcolor{Goldenrod}{95.5} & \textcolor{orange}{83.5} & \textbf{96.4} \\
        \hline
    \end{tabular}
    \end{adjustbox}
    \vspace{0.05cm}
    \caption{$Recall@5$ of some of the SOTA foundation-model-based VPR methods on various test datasets. All methods are trained on the most diverse VPR training dataset: the GSV-Cities dataset. The second row represents the domain gap of the respective test dataset from the GSV-Cities training dataset. $\checkmark$ indicates a small gap and $\checkmark \checkmark \checkmark$ indicates a large gap. On average, BoQ-DinoV2 is the SOTA in VPR, outlined in Bold, and thus our primary baseline. To indicate the margin of improvement left for BoQ, the datasets are ranked from left to right and colored. Datasets with a small train-test gap are almost solved, but a large train-test domain gap presents a challenge even for the SOTA VPR methods.}
    \label{tab:vprontestdatasets}
\end{table*}


\subsection{Our proposed Reference-Set-Finetuning (RSF)}
\label{sec:finetuningG}
\begin{figure}
    \centering
    \includegraphics[width=1.0\linewidth]{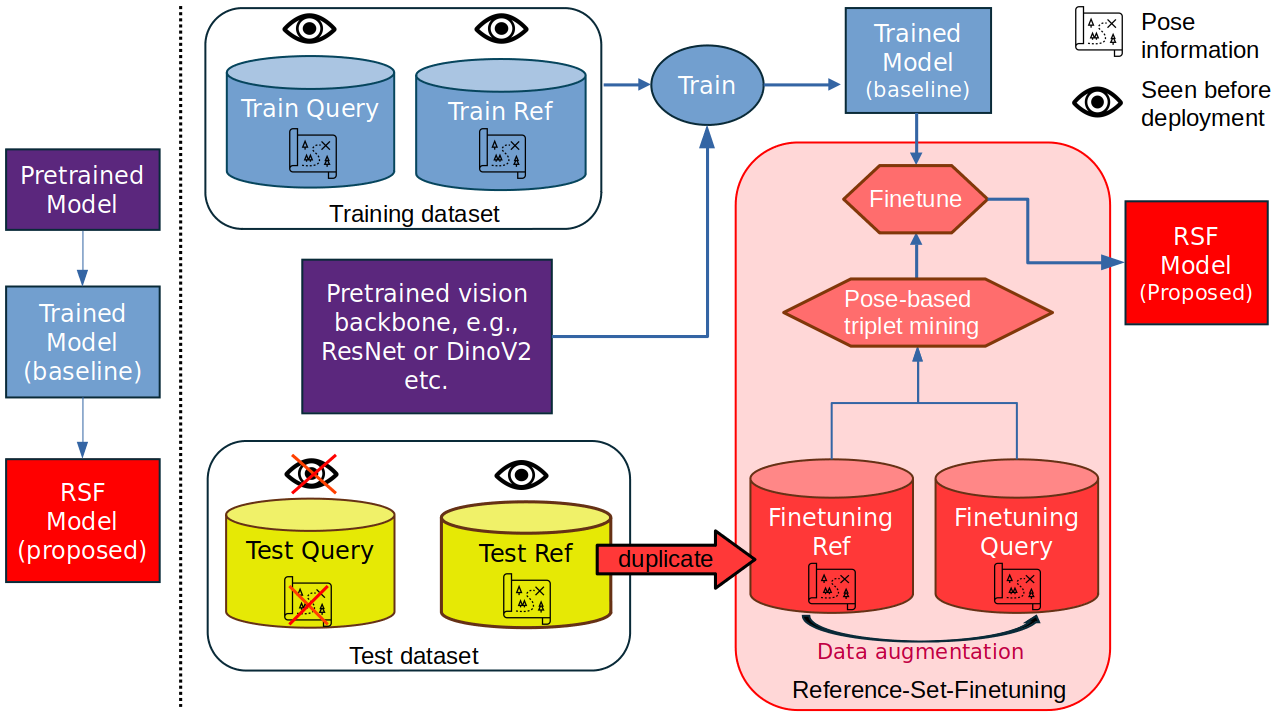}
    \caption{Deep learning for VPR usually utilizes a pretrained neural network that is further trained on a VPR dataset in a supervised manner with ground-truth poses. This usual pipeline assumes that we do not have any access to the test environment and that the training dataset is diverse enough to cover features of the test domain. However, there is always a train-test domain gap. We propose that the reference images in the test set are freely available offline in VPR and could be used to finetune VPR methods using simple data augmentations. This novel take on the problem setting of VPR, results in reference-set-finetuned (RSF) models that are more robust than the original trained model.}
    \label{fig:rsf_scheme}
\end{figure}

The preceding discussion suggests that although the training dataset $D$ could be carefully curated to maximize diversity, it might still lack the domain knowledge needed for $G$ to perform well on the test-time queries $\setImQ$. 
Here we make our key observation: $\setImRef$ is already available at the map preparation stage as well as its corresponding set of poses $\setPoseRef$. Therefore, we propose \textit{Reference-set-finetuning (RSF)}, an unexplored but straightforward and effective procedure to adapt a trained model $G$ to the target domain. Concretely, RSF (1) creates a \textbf{finetuning dataset} $D_{ft}=(\setImQ^{ft}, \setPoseQ^{ft}, \setImRef^{ft}, \setPoseRef^{ft})$, and (2) updates $G$ on $D_{ft}$ with pose-aware triplet mining, as illustrated in Fig.~\ref{fig:rsf_scheme}, and described in the following.

For $D_{ft}$, the finetuning query set $\setImQ^{ft}$ should represent a combination of viewpoint and appearance changes typically seen between the matching queries and references. Thus, a query $I_q^{ft} \in \setImQ^{ft}$ is formulated as $I_q^{ft} = A(I_i^{ft})$, where $A(.)$ represents an \textbf{augmentation operation}. Ideally, $A(.)$ approximates the viewpoint and appearance changes expected between the queries and references. An $M$ number of different augmentations could be chosen as $A(.)$. In conclusion, the choices follow: 
\begin{align}    
\setImRef^{ft} = \setImRef, \qquad \\
\setPoseRef^{ft} = \setPoseQ^{ft}  = \setPoseRef, \qquad \\
\textrm{and} \qquad |\setImQ^{ft}| = M \times |\setImRef^{ft}|. 
\end{align}

The finetuning queries $\setImQ^{ft}$ and references $\setImRef^{ft}$ are encoded as feature vectors with $G$, positives and hard negatives~\cite{arandjelovic2016netvlad} are \textbf{mined given the poses} $\setPoseQ^{ft}$ and $\setPoseRef^{ft}$, and the network $G$ is \textbf{finetuned} using a standard triplet loss~\cite{gordo2017end}:
\begin{align}
L_{triplet} = max\{d(f_q^{ft}, f_p^{ft})-d(f_q^{ft}, f_n^{ft})+m,0\},    
\end{align}

with a Euclidean distance function $d(f_1,f_2)=||f_1-f_2||_2$ and a margin $m$. A hard-negative for a given query is the wrong reference image further than some fixed physical distance threshold that is the closest in the feature space.



%% file: chapters/experiments.tex
\section{Experiments}
First, we present the experimental setup of our work, then report the qualitative and quantitative performance of RSF models compared to baselines, and finally evaluate the various aspects of RSF. 


\subsection{Datasets and evaluation metric}
To evaluate RSF, we use three public VPR datasets which have a large train-test domain gap and hence pose challenges to SOTA VPR methods, and one dataset with a small train-test domain gap.
Our ground-truth usage is similar to the standard formats in VPR~\cite{berton2022deep}, 
All of these datasets are summarized in Table~\ref{tab:datasetssummary}. 

\iccvcolor{The \textbf{Nordland dataset}~\cite{nordlanddataset} consists of a railway-track traversal through Norway during two different seasons: summer and winter. The summer traversal acts as reference images while the winter images are queries. This dataset is challenging due to the unstructured environment depicted in different seasons. We also use the challenging AmsterTime dataset~\cite{yildiz2022amstertime} that contains archival imagery of Amsterdam and its corresponding Google Street View images. We use the archival images as references and street view images as queries, which depicts the task of retrieving an archival image of a place given a query image. We refer to this version as \textbf{AmsterTime-AR dataset}, outlining that the \underline{A}rchival images acts as \underline{R}eferences. We use the \textbf{Eynsham dataset}~\cite{cummins2009highly} that contains only grayscale images presenting a lack of color information for VPR. Finally, we use the \textbf{SVOX-Night dataset}~\cite{berton2021adaptive} that contains night-time images as queries and day-time images as references collected through Google Street View (GSV) in Oxford.}

\begin{table}
    \centering
    \begin{tabular}{c|ccccc}
         & Queries & Refs.  & Q-R gap & Train-test gap\\
         \hline
        Nord. & 27.6k  & 27.6k & $\checkmark \checkmark$ & $\checkmark \checkmark \checkmark$ \\
        Amst-AR & 1231 & 1231 & $\checkmark \checkmark \checkmark$  & $\checkmark \checkmark \checkmark$  \\
        Eyns. & 24k & 24k & $\checkmark$  & $\checkmark$ $\checkmark$  \\
        SVOX-Ni & 823 & 17.2k & $\checkmark \checkmark$  & $\checkmark$\\

    \end{tabular}
    \caption{The datasets used in this work. We report the total number of query images, the total number of reference images, the presence of a domain gap between the queries and references, and the presence of a domain gap between the respective test dataset and the GSV-Cities training dataset. $\checkmark$ indicates a small gap and $\checkmark \checkmark \checkmark$ indicates a large gap.}
    \label{tab:datasetssummary}
    \vspace{-0.7cm}
\end{table}

Following the existing literature, Recall@N is used as the evaluation metric. 
Ground-truths are as-is used by others~\cite{lu2024cricavpr, ali2024boq, berton2022deep, izquierdo2024optimal}. 
A retrieval is successful if the Top-N retrieved reference images were within a 25-meter radius of the query image.

\subsection{Implementation details}
\begin{figure}
    \centering
    \includegraphics[width=1.0\linewidth]{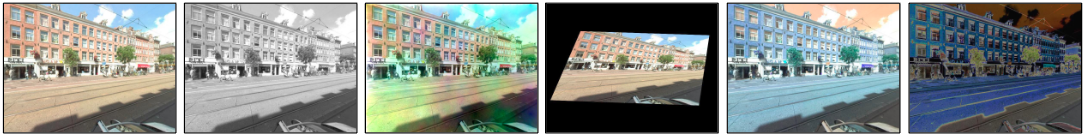}
    \caption{Examples of the augmentations applied to create finetuning queries using Kornia augmentations~\cite{riba2020kornia}. Left-most is the original reference image.}
    \label{fig:kornia_augmentedimages}
    \vspace{-0.35cm}
\end{figure}

Given the standards and SOTA described earlier in section~\ref{sec:standardvprformulation}, Dino-V2~\cite{oquab2024dinov2} backbone with BoQ~\cite{ali2024boq} aggregation trained on the GSV-cities dataset is used as the primary baseline VPR method $G$, since it is the current SOTA in VPR. Nevertheless, we also report the performance of SALAD~\cite{izquierdo2024optimal} when used with the proposed RSF. We use the complete reference set of each respective test dataset for performing RSF as described in section~\ref{sec:finetuningG}. A small learning rate of 1e-7 is used for all datasets for both the VPR techniques. Simple image-level augmentations from the Kornia library~\cite{riba2020kornia} are used as $A$; examples are shown in Fig.~\ref{fig:kornia_augmentedimages}. More sophisticated augmentations, such as domain translations using image-to-image vision foundation models, could also be considered~\cite{brooks2023instructpix2pix}. The Kornia augmentations are applied on the fly and randomly chosen during training. To avoid overfitting the test set, we validate our model on the Pitts30k validation set~\cite{berton2022deep}.
RSF is done on a single NVIDIA A100 80GB GPU and, on average, takes only a few hours ($\approx 1-5$) depending on the size of the reference set and the method $G$.

\subsection{Results}

\begin{table*}
    \centering
    \begin{adjustbox}{width=\textwidth}
    \begin{tabular}{ccccccccc|cc} \\
         & \multicolumn{2}{c}{Nordland}  &  \multicolumn{2}{c}{Amster-AR} & \multicolumn{2}{c}{SVOX-Night} & \multicolumn{2}{c}{Eynsham} & \multicolumn{2}{c}{Average} \\
           & $R@1$ & $R@5$  & $R@1$ & $R@5$ & $R@1$ & $R@5$ & $R@1$ & $R@5$ & $R@1$ & $R@5$ \\
        \hline
        MixVPR~\cite{ali2023mixvpr}  & 76.1 & 86.8 & 38.3 & 60.4 & 63.1 & 79.5 & 89.4 & 93.2 & 66.7 & 80.0 \\
        BoQ-Res~\cite{ali2023mixvpr}  & 83.3 & 91.1 & 52.1 & 75.4 & 85.7 & 94.7 & 91.2 & 94.9 & 78.1 & 89.0 \\
        CricaVPR~\cite{lu2024cricavpr}  & 91.2 & 96.2 &  64.7 & 83.9 & 86.9 & 95.0 & 91.6 & 94.9 & 83.6 & 92.5 \\
        SALAD~\cite{izquierdo2024optimal}  & 85.9 & 93.5 &  58.7 & 79.7 & 95.0 & 99.3 & 91.5 & 95.0 & 82.8 & 91.9 \\
        BoQ~\cite{ali2024boq}   & 90.4 & 95.9 &  61.9 & 83.5 & 97.1 & 99.4 & 92.1 & \textbf{95.5} & 85.4 & 93.6 \\
        \hline
        SALAD-RSF & 91.4 & 96.2 & 59.9 & 80.6 &	96.1 & 98.8 & 91.8 & 95.2 &	84.8 & 92.7\\
        BoQ-RSF  &  \textbf{94.2} & \textbf{97.7} &  \textbf{65.6} & \textbf{86.3} & \textbf{98.8} & \textbf{99.6} & \textbf{92.2} &  95.4 & \textbf{87.7} & \textbf{94.8}\\
    \end{tabular}
   \end{adjustbox}
   \vspace{0.1cm}
    \caption{\iccvcolor{The recalls of SOTA VPR methods tested on various challenging test datasets. The first two rows: MixVPR and BoQ-Res use ResNet-50 backbone, while the remainder use DinoV2 backbone. All methods are trained on the GSV-Cities dataset. Best is in Bold.}}
    \label{tab:performance_rsf}
\end{table*}

\textbf{Baseline comparison:}
\iccvcolor{Table~\ref{tab:performance_rsf} contains the performance of RSF models in comparison to baselines. Models finetuned using our proposed RSF outperform existing methods by a large margin for both metrics. Please note that this performance improvement is \textit{without} the use of new training data or a stronger backbone. The performance benefits are more significant for the challenging Nordland and AmsterTime-AR datasets, which are the primary focus due to their large train-test domain gap. We also note that the proposed RSF is beneficial for the datasets without a large train-test domain gap, e.g., the SVOX-Night and Eynsham datasets. However, the performance improvement is less significant than on other datasets. More importantly, we show that both the SOTA VPR methods, BoQ and SALAD, benefit from RSF.}

\begin{figure*}[h]
    \centering
    \includegraphics[width=0.95\linewidth]{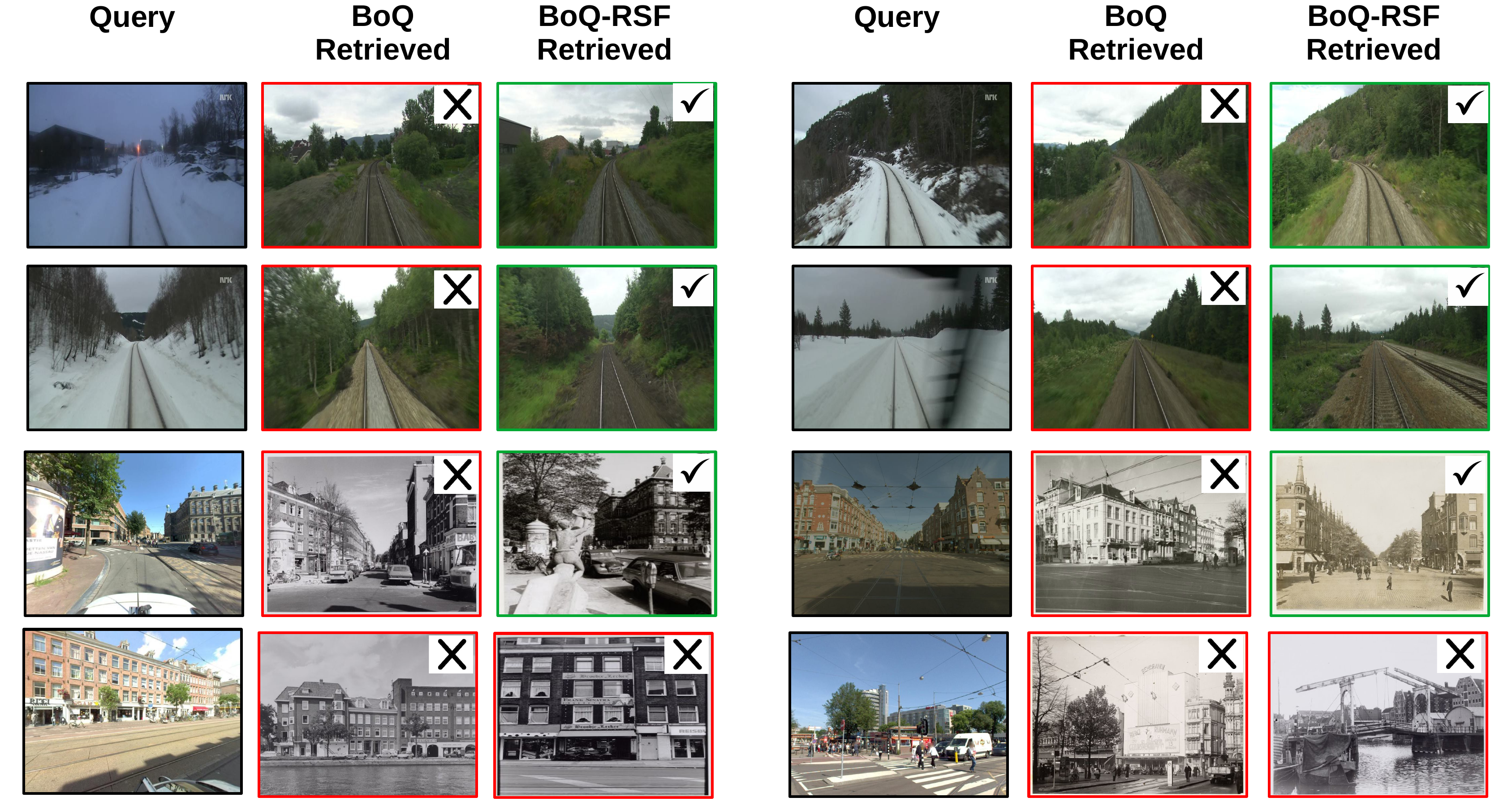}
    \caption{Examples of queries that are mismatched by the original BoQ-DinoV2 model but correctly matched by our reference-set-finetuned BoQ-RSF model, except for the last row which demonstrates two BoQ-RSF failure cases.}
    \label{fig:boqrsf_exemplar_images}
\end{figure*}

We further show in Fig.~\ref{fig:boqrsf_exemplar_images} examples of queries that are correctly matched after the proposed RSF, and also some failure cases. \iccvcolor{Since BoQ with RSF is the best-performing method in our baseline comparison, we focus on this method in the remainder of the experiments.}

\textbf{Model generalization:} A key component of this study is the desire for the RSF models to retain generalization to the other test datasets. For this, we report in Table~\ref{tab:generalization} the performance of an RSF model finetuned on a given reference dataset and evaluated on the other test datasets. Interestingly, we note that not only do the finetuned models retain generalization to other test datasets, but also that the RSF finetuned models consistently outperform the original model, agnostic to the reference set used for finetuning. This is attributed to the additional finetuning of SOTA on VPR-specific data; however, quite expectedly, we see a diagonal trend in the bold numbers, such that the best-performing RSF model for each test dataset is always the model finetuned on the same test dataset's reference map. 


\begin{table}
    \centering
    \begin{tabular}{c|ccc}
         & \multicolumn{3}{c}{\textbf{Test dataset}} \\
         & Nord. &  Amst-AR & SVOX-Ni. \\
        \hline
        Baseline BoQ & 90.4 & 61.9 & 97.1 \\
        \hline
        BoQ-RSF (Nord.) & \textbf{94.2} & 64.4 & \textbf{98.9}  \\
        BoQ-RSF (Amst-AR) & 92.3 & \textbf{65.6} & \textbf{98.9} \\
        BoQ-RSF (SVOX-Ni.) & 93.4 & 64.7 & \textbf{98.9} \\
    \end{tabular}
    \vspace{0.1cm}
    \caption{The Recall@1 of RSF models on various test datasets. The first column reports the reference set used for BoQ-RSF. RSF models retains generalization. Bold numbers in the diagonal indicate that the best-performing method for each dataset is the model finetuned on that dataset's reference set.}
    \label{tab:generalization}
\end{table}

\begin{figure*}
    \centering
    \includegraphics[trim={0 5.2cm 0 0}, clip, width=0.95\linewidth]{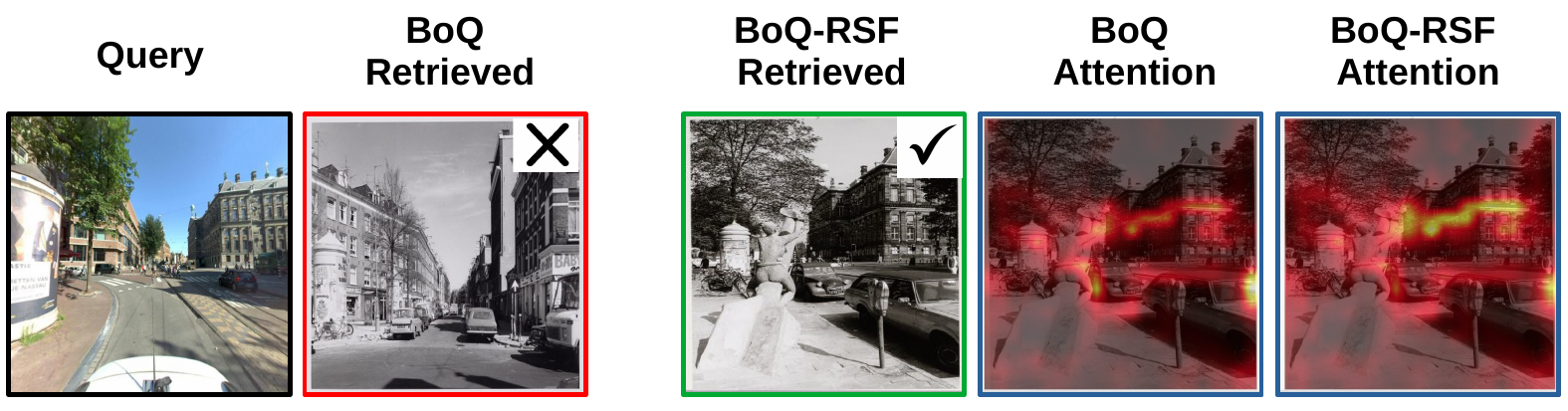}
    \caption{Learned attention for the original BoQ and the BoQ-RSF model on a ground-truth reference image is shown. The RSF model attends more to facades in the building while BoQ attends to edges. These attention masks are for the \textit{same} BoQ query of the original and the BoQ-RSF model.} 
    \label{fig:boqrsf_attentionmasks}
\end{figure*}

\textbf{Attention masks:} We visualize the attention masks for a learned BoQ query in Fig.~\ref{fig:boqrsf_attentionmasks} for the original model and the RSF model. Note that the RSF model strongly attends to the unique facades of windows in the building on the right, while the original BoQ only attends to edges. 


\subsection{Ablations}
We have argued in this work that the reference poses are freely available offline in VPR and are thus used in pose-based triplet mining for RSF. However, it is possible to have image-retrieval use-cases where reference images are available without pose information, e.g., image cataloging, landmark identification, etc. Table~\ref{tab:rsfwithandwithoutposes} thus reports the performance of our baseline in comparison to RSF models trained with and without access to pose information in the reference set. It is observed that although the reference pose information is helpful for RSF and such models are consistently the best-performing, but even without access to reference pose information, RSF models are still better than the baseline. 

\begin{table}[h]
    \centering
    \begin{tabular}{c|ccc}
                 & Nordland  & Amst-AR \\
                 \hline
       Baseline BoQ  &  95.9	& 83.5 \\
       BoQ-RSF (without poses)  & 97.1	& 85.3 \\
       BoQ-RSF (with poses) & \textbf{97.7}	& \textbf{86.3} \\
    \end{tabular}
    \caption{The Recall@5 performance of a baseline BoQ method is compared with RSF two test datasets with and without access to the test-time reference poses. The availability of test-time reference poses allows for hard-negative mining and gives SOTA performance compared to random negative mining when pose information is not accessible. However, even without access to the reference poses, RSF model performs better than the baseline BoQ.}
    \label{tab:rsfwithandwithoutposes}
\end{table}

We further report in Table~\ref{tab:rsfwithsiffaugmentations} the effect of Kornia augmentations on our proposed RSF for BoQ. These results show that augmentations are required to benefit from fine-tuning on the reference set, and that appearance augmentations are more useful than viewpoint augmentations for the chosen datasets. Only having viewpoint augmentations and no appearance augmentations is hurtful for RSF. We hypothesize that using viewpoint augmentations as $A$ is distractful for the model finetuned on the Nordland dataset, since there is almost no viewpoint change between the queries and the references in this dataset. The choice of augmentations in practice should follow from the expected query-reference domain gap, and in case of no prior knowledge about the expected  Q-R gap, we recommend that the viewpoint augmentations be used together with appearance augmentations as a thumb rule.

\begin{table}[h]
    \centering
    \begin{tabular}{c|ccc}
             Chosen $A$     & Amster-AR & Nordland \\
                 \hline
No augmentations	& 83.51 & 95.92	 \\
\hline
No viewpoint augmentations		& \underline{86.31} & \textbf{97.80}	 \\
No appearance augmentations		& 76.20  & 91.13\\
All augmentations	& \textbf{86.32}  & \underline{97.70}	\\
    \end{tabular} 
    \caption{The Recall@5 performance of BoQ-RSF with different types of augmentations chosen as $A$.}
    \label{tab:rsfwithsiffaugmentations}
\end{table}


%% file: chapters/conclusions.tex
\section{Conclusions}
In this work, we demonstrate that even the strong vision-foundation models-based VPR methods trained on large-scale Google Street View data struggle on test datasets that represent a domain different from the training data. We thus proposed that the reference set in test datasets is a free and valuable source of information that can be used to bridge this train-test domain gap. A simple Reference-Set-Finetuning (RSF) strategy is proposed that boosts the performance of SOTA VPR methods by large margins. The proposed RSF is shown to work for multiple datasets. The resulting finetuned models retain generalization to other test datasets. We also show that the same RSF strategy could be applied to other VPR methods, albeit the performance benefits vary. Future works could investigate further how different formulations of RSF, particularly the augmentations, could benefit different VPR methods. 

\textbf{Acknowledgments}: This work was funded by the TU Delft AI initiative through the 3D Urban Understanding Lab (3DUU) at TU Delft. Part of the work was conducted during Mubariz Zaffar's research exchange visit to the AIRLab at Carnegie Mellon University, headed by Dr. Sebastian Scherer. This exchange was supported by the North-America-Travel-Grant from the Delft University Fund. 